[1994/12/01]
\documentclass[leqno,fleqn,11pt]{article}
\usepackage[margin=0.75in]{geometry}  
\usepackage[largesmallcaps]{kpfonts}
\usepackage{microtype}
\usepackage{amsmath,amsthm}
\usepackage{xy}		
\usepackage{graphicx}
\usepackage{caption}
\usepackage{subcaption}
\usepackage[dvipsnames]{xcolor}

\usepackage[round,sort&compress]{natbib}

\usepackage{amsmath}   
\usepackage{xcolor}
\usepackage{hyperref}
\usepackage{alphalph}
\usepackage{dsfont}   
\usepackage{etoolbox}
\usepackage{marginnote}				
\usepackage{datetime}
\reversemarginpar

\renewcommand{\sc}[1]{{\textsc{#1}}}
\renewcommand{\sf}[1]{{\color{blue}\textsf{#1}\color{black}}} 
\renewcommand{\vec}[1]{\overrightarrow{\mathsf{#1}}}
\newcommand{\gap}{\vspace{2mm}\noindent} 

\newcommand{\TPT}{NECSTransformer}

\newcommand{\un}{\mbox{$\sf{un}$}}
\newcommand{\lock}{\mbox{$\sf{lock}$}}
\newcommand{\able}{\mbox{$\sf{able}$}}
\newcommand{\lockable}{\mbox{\sf{$[ \lock \ \able ]$}}}
\newcommand{\unlock}{\mbox{\sf{$[ \un \ \lock ]$}}}

\newcommand{\Lr}{\mbox{\textit{\sf{L}}}}
\newcommand{\Rr}{\mbox{\textit{\sf{R}}}}

\newcommand{\Nfont}[1]{\mbox{\color{blue}$\mathbf{\mathtt{#1}}$\color{black}}}
\newcommand{\NinN}{\Nfont{n}-in-\textit{n}}

\newcommand{\squeeze}[2]{\scalebox{#1}[1.0]{#2}}

\newcommand{\scisec}[1]{\vspace{5mm}\noindent{\color{BrickRed}\large \textbf{\textsf{#1}}}\vspace{3mm}\newline\noindent} 
\newcommand{\scisubsec}[1]{\vspace{3mm}\noindent{\color{BrickRed} \textbf{\textsf{#1}}}}

\newcommand{\eA}[1]{{\textit{#1}}}   
\newcommand{\eB}[1]{{\textit{#1}}} 
\newcommand{\bi}[1]{{\textit{\textbf{#1}}}}
\newcommand{\e}[1]{{#1}}   
\newcommand{\Gi}{1G}
\newcommand{\Gii}{2G}
\newcommand{\Giii}{3G}

\title{Neurocompositional computing:\\ 
From the Central Paradox of Cognition \\
to a new generation of AI systems}
 
\date{}
\author
{Paul Smolensky,$^{1,2\ast}$  R. Thomas McCoy,$^{1\ast}$ Roland Fernandez,$^{2}$ \\
Matthew Goldrick,$^{3}$ Jianfeng Gao$^{2}$\\
\\
\normalsize{$^{1}$Department of Cognitive Science, Johns Hopkins University}\\
\normalsize{3400 N. Charles St., Baltimore, MD 21218, USA}\\
\normalsize{$^{2}$Microsoft Research; Redmond, WA 98052, USA.}\\
\normalsize{$^{3}$Department of Linguistics, Northwestern University; Evanston, IL 60208, USA.}\\
\\
\footnotesize{$^\ast$To whom correspondence should be addressed; Email:  \sf{\{paul.smolensky,tom.mccoy\}@jhu.edu.}}
}

\begin{document} 

\maketitle 

\begin{abstract} 
What explains the dramatic progress from 20th-century to 21st-century AI, and how can the remaining limitations of current AI be overcome? 
The widely accepted narrative attributes this progress to massive increases in the quantity of computational and data resources available to support statistical learning in deep artificial neural networks. 
We show that an additional crucial factor is the development of a new type of computation.
\textit{Neurocompositional computing} \citep{smolensky2022neurocompositional} adopts two principles that must be simultaneously respected to enable human-level cognition: the principles of Compositionality and Continuity. 
These have seemed irreconcilable until the recent mathematical discovery that compositionality can be realized not only through discrete methods of symbolic computing, but also through novel forms of continuous neural computing. 
The revolutionary recent progress in AI has resulted from the use of limited forms of neurocompositional computing. 
New, deeper forms of neurocompositional computing create AI systems that are more robust, accurate, and comprehensible.
\end{abstract}

\filbreak

\noindent
Artificial Intelligence (AI) has been a long time coming of age.
It was Ada Lovelace who  recognized in 1843 the profound implications, extending far beyond numerical calculation, of Charles Babbage's design for a general-purpose numerical computer:
“in enabling mechanism to combine together \textit{general} symbols, in successions of unlimited variety and extent, a uniting link is established between the operations of matter and \textellipsis\ abstract mental processes” \cite[Note A, p. 368]{bowden1953faster}. 
Now, more than a century and a half later, the widespread deployment of AI has become a landmark achievement of the 21st century, a turning point that has been likened to the industrial revolution of the 18th century \citep{Thomas2020NewIndustrialRevolution}.
Properly understanding this breakthrough is critical for controlling how AI develops in the remainder of the century, both for maximizing its benefits and minimizing its hazards \citep{grosz2018century}.
Here we present a new analysis of this progress, an analysis which is yielding a new generation of AI systems. 

The recent progress in AI---specifically, AI based on artificial neural networks exploiting deep learning---has typically been attributed to merely quantitative technological improvements: greatly increased computing power and data quantity \citep[p. ix]{sejnowski2018deep}.
Correct as far as it goes, this view overlooks a major factor, namely the advent of \textit{neurocompositional computing}, which, according to a contemporary theory in cognitive science \citep{smolensky2006harmonic}, is the form of computing underlying human intelligence. 
We argue that, in addition to the critical expansion of computing resources, it is the unrecognized emergence of neurocompositional computing---albeit only implicitly, in an acutely restricted form---that has powered the most dramatic recent advances in AI.

Despite the unprecedented progress, it is generally recognized that contemporary AI systems still suffer serious deficiencies, including extreme opacity and weakness in learning general knowledge from limited experience.
Here, we show how these problems are being addressed through a deeper development of neurocompositional computing, deployed in the new AI systems we review.
In addition to learning more robustly, these new systems are more comprehensible than standard deep learning networks, diminishing the opacity which limits both their utility \citep{Turek2016XAI} and our ability to control their sometimes societally-deleterious behaviors \citep{bender2021dangers}.

The original contributions of this article, presented in successive sections, are these:

\vspace{-2mm}
\begin{itemize}
    \item identifying key contributors to the power of human intelligence---the Continuity and Compositionality Principles---which together define neurocompositional computing;
    \item diagnosing the limitations of 20th-century AI as rooted in the use of computing architectures that failed to respect both these principles;
    \item elucidating the recent progress in AI as resulting in significant part from simultaneous adherence to both principles---achieving neurocompositional computing to a limited degree;
    \item presenting a general formalism from cognitive science that allows considerably deeper realization of neurocompositional computing: Neurally-Encoded Composi\-tionally-Struc\-tured Tensor (NECST) computing;
    \item showing how new NECST AI models harness the power of neurocompositional computing to mitigate limitations of previous AI systems;
    \item illustrating how NECST computing substantially improves both the generalization \squeeze{0.91}{capacity and the comprehensibility---and consequently the controllability---of AI systems.}
\end{itemize}

\scisec{The Central Paradox of Cognition and the principles defining \\ neurocompositional computing}
What type of computing enables human intelligence?
Because human intelligence arises from the brain, the obvious answer is: \e{neural computing} \citep{churchland2016computational},
in which information is encoded in numerical \e{activation vectors}---patterns of activation over groups or layers of neurons  (Fig.~1\textit{a}). 
These encodings form a vector space; the coordinates of an encoding in this space are the numerical activation values of the neurons hosting the activation pattern (Fig.~1\textit{b}).
The activation vector that encodes an output results from spreading the activation that encodes an input among multiple layers of neurons through connections of varying strengths or weights.
In a typical neural network AI model, the values of these weights are set by training the model on examples of correct input/output pairs. 
After seeing many such pairs, the model will ideally converge to connection weights that produce the correct output when given an input, not only for the inputs seen during training, but also for novel inputs in a test set.

\begin{figure}[!ht]
    \centering
    \includegraphics[width=\textwidth]{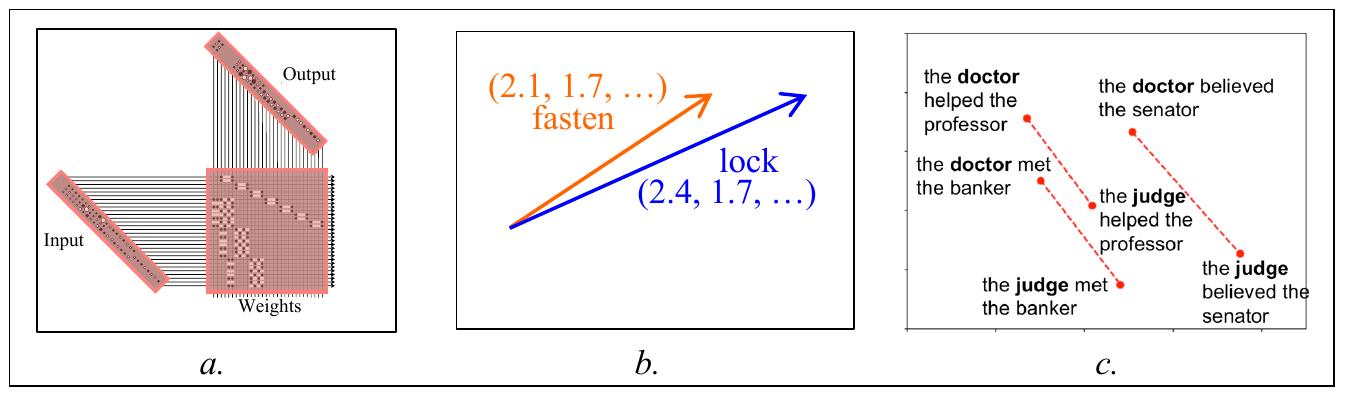}
    \centering
    \caption{Neural computing---Encoding information in numerical vector spaces.}
    \label{fig:1A}
\end{figure}

Once a computational task is encoded in numerical vectors, solving it becomes a problem in statistical inference.
A neural network is a complex statistical model with many parameters that are estimated from the statistics of the training data. During learning, the model can compute, and store in its weights, the relations between the numbers in the vector encodings produced when processing the training examples; after learning, during testing, it can use these learned relations to infer an output vector given an input vector that encodes a novel input.

Neural computing respects the \eA{Continuity Principle:} the encoding and processing of information is formalized with real numbers that vary continuously, i.e., by amounts that can be arbitrarily small.
Continuity means that knowledge about information encoded in one vector automatically generalizes to similar information encoded in nearby vectors \citep{hinton1986distributed}. 
For instance, standard neural systems can readily generalize their knowledge about the word \textit{lock} to the word \textit{fasten} (Fig.~1\textit{b}): these words appear in similar contexts within texts used for training (they have similar `distributional meaning'), so neural-network learning algorithms will encode them as vectors that are close together; as a result, neural processing of these two encodings will produce similar results---similarity-based generalization.

The structure of the vector space can capture other types of relationships as well, such as having a consistent offset that represents a systematic difference between pairs of sentences (Fig.~1\textit{c}).
Most importantly, continuity enables \e{deep learning}, in which a model's connection weights---even those deeply buried within the network, far from the input and output encodings---can be modified gradually to smoothly improve the model's statistical inference of outputs from inputs in its training set \citep{rumelhart1986learning}.

Although it appears obvious that cognition deploys neural computing, virtually all aspects of human intelligence have been formally and precisely described since antiquity \citep{kiparsky1969syntactic} in terms of a very different type of computing: \e{compositional-structure processing} \citep{janssen2012CompHistory}.
In this type of computing (Fig.~2), complex information is encoded in large structures---compositional encodings---which are built by composing together smaller structures that encode simpler information.
To process the complex information encoded in a large structure, it suffices to compose together the results of processing the simpler information encoded in its smaller substructures. 
This is the \eA{Compositionality Principle} \citep{szabo2012case}.

Compositionality has long been seen as key to the power of human cognition: it enables strong \eA{compositional generalization}, enabling us to understand any one of a potentially infinite number of novel situations by encoding the situation internally as a novel composition of familiar, simpler parts, and composing together our understanding of those parts.  
In virtually all domains of cognition---from vision and speech to reasoning and planning---empirical investigation over millennia has consistently shown how cognitive functions can be well approximated as computation over appropriate compositional structures, structures with respect to which human cognition exhibits strong compositional generalization (\citealt{hinzen2012oxford}).
The parts which compose to form such encodings are sometimes familiar elements like objects, words, concepts, and actions, and sometimes scientifically-discovered units like phonemes---the basic sounds of speech which combine to form words, each of which is typically denoted by a single letter in alphabetic writing systems.  

Robust compositional generalization is all-pervasive in cognition: it underlies the power of both fast, automatic, intuitive, largely unconscious cognition (e.g., visual processing of novel arrangements of familiar objects), as well as slow, controlled, deliberative, conscious cognition (e.g., playing chess with a novel initial arrangement of pieces) \citep{kahneman2011thinking}.

In sum: human cognition gets tremendous power by understanding that \textit{the world is strongly compositional}.

\gap \noindent
Compositional encodings have for centuries been formalized as intricate arrangements of symbols---\e{symbol structures} like those we use to represent expressions in algebra or formal logic \citep{newell1980physical}.
To take a simple example, the two symbol structures \sf{$[ \un \ [ \sf{lock} \ \sf{able} ] ]$} and \sf{$[ [ \sf{un} \ \sf{lock} ] \ \sf{able} ]$} formalize compositional word encodings that correspond to the two meanings of \textit{unlockable}: not able to be locked (\textit{un-lockable}), and able to be unlocked (\textit{unlock-able}), respectively.

Fig.~2 illustrates symbolic compositional-structure processing with three functions (notated `input $\mapsto$ output').
In the domain of language, Fig.~2\textit{a} shows the Tree Adjoining function, which here takes as input the phrase-structure trees for the sentence \sf{Kim hates symbols} and the adverb \sf{really} and inserts the latter into the middle of the former to produce as output the phrase-structure tree for \sf{Kim really hates symbols}.
This is an operation that is known to give grammars sufficient power to achieve the level of complexity displayed by the syntax of sentences in human languages \citep{joshi1985tree}.
In the domain of mathematics, Fig.~2\textit{b} shows a function that simplifies a ratio of ratios into a product of ratios.
In the domain of logical reasoning, Fig.~2\textit{c} shows a function that chains if-then implications: it reads \textit{q implies r and r implies s entails q implies s}, e.g., knowing that \textit{if it rains then I will drive} ($q \rightarrow r$) and \textit{if I drive then I will need to charge my car} ($r \rightarrow s$) allows us to conclude \textit{if it rains then I will need to charge my car} ($q \rightarrow s$).

\begin{figure}[!ht]
    \centering
    \includegraphics[width=\textwidth]{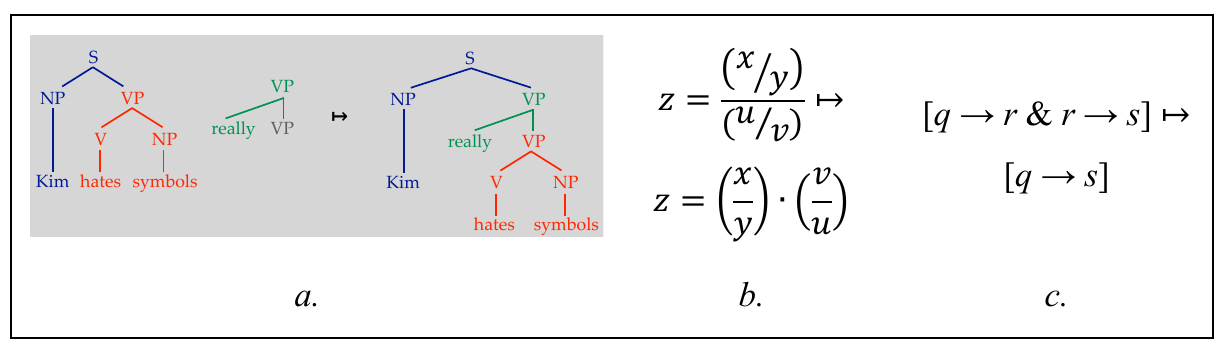}  
    \centering
    \caption{Symbolic compositional-structure computing---Encoding information in symbol structures.}
    \label{fig:1B}
\end{figure}

The symbols in these structures are abstract entities that are by definition inherently discrete: unlike real numerical values, they cannot be modified by arbitrarily small amounts (although physical drawings of what we perceive as symbols can be).
The identity of a symbol, and its position in a symbol structure, are all-or-none: a symbol either is an \sf{A} or it isn't; it either occupies the first position of a symbol sequence or it doesn't.

As the contrast between Figs.~1 and 2 vividly illustrates, discrete symbolic composition\-al-structure-processing computers and continuous neural computers are profoundly different.
Yet, somehow, the computer in our heads apparently is simultaneously a neural computer and a composition\-al-structure computer \citep{marcus2001algebraic}. 
How can this be?
We call this the \e{Central Paradox of Cognition} \citep{smolensky1988proper}.

One tempting approach for resolving this paradox would be to posit a hybrid system that combines separate symbolic and neural machines \citep{andreas2016neural,wang2020rat,marcus2019rebooting}. 
However, such an approach is only viable if the various components of cognition can be cleanly separated into those that are compositional and those that are continuous. As discussed below, this is not the case: in central cognitive domains, individual encodings respect both the Compositionality and the Continuity Principles. 
Therefore, a true resolution of the Central Paradox requires a new type of computing---\e{neurocompositional computing}---that \eB{simultaneously} satisfies both the Continuity and Compositionality Principles.

\vspace{4mm}
\scisec{20th-century AI limitations: Violating the principles}
Evidence that human intelligence requires simultaneously respecting both the Continuity and Compositionality Principles comes from a perhaps surprising place: 20th-century AI.

Despite impressive advancements that inspired recurrent optimism, the AI systems built on symbolic computing have consistently fallen far short of human general intelligence \citep{minsky1991logical}.
The discrete material of symbolic encodings---e.g., symbols denoting words---have often proved overly rigid to meet the subtle demands of human cognition; for instance, while such systems can easily generalize from \textit{lock} and \textit{-able} to \textit{lockable}, they cannot easily generalize from \textit{lock} to \textit{fasten} because those two words must be encoded as entirely independent discrete symbols.
And the discreteness of the structures themselves also made them insufficiently flexible: even a simple discrete sequential structure for the basic sounds of speech---roughly corresponding to individual letters---turns out to be too rigid, as some sounds cannot be pinned down to a single sequential position but rather simultaneously occupy, to continuously varying degrees, a blend of multiple positions \citep{smolensky2016gradient}. 
The strengths and weaknesses of AI systems deploying symbolic computing (Fig.~3\textit{e--h}) arise from these systems respecting the Compositionality Principle, but not the Continuity Principle.

The AI systems (and cognitive models) of the 20th century that employed neural computing typically did the reverse: they respected the Continuity Principle but violated the Compositionality Principle.
Continuous vector processing endowed these models with the power to approximate arbitrarily closely any vector-to-vector function \citep{hornik1989multilayer}; this suffices for tasks that do not demand a high degree of compositional processing, such as classifying the texture of a single image patch.
But violating the Compositionality Principle brought serious limitations which were prominently emphasized by Jerry Fodor and Zenon Pylyshyn (\citeyear{fodor1988connectionism}) as well as Gary Marcus (\citeyear{marcus2001algebraic}).
Although neural networks are far less rigid than their symbolic counterparts (Fig.~3\textit{a--b}), they typically suffer from complementary weaknesses (Fig.~3\textit{c--d}), including striking failures of compositional generalization \citep{kim2020cogs,baroni2020linguistic}.
For instance, as noted above, standard neural systems can easily generalize their knowledge about \textit{lock} to \textit{fasten}, 
but these networks cannot readily generalize to novel compositional structures involving \textit{lock}, such as \textit{lockable}: they lack the compositional encodings that support such generalization. 

In sum: traditional neural AI systems suffer because \textit{they do not understand that the world is strongly compositional}---they can't, because they lack the compositional encodings needed to realize such understanding.

\vspace{4mm}
\begin{figure}[!ht]
    \centering
    \includegraphics[width=\textwidth]{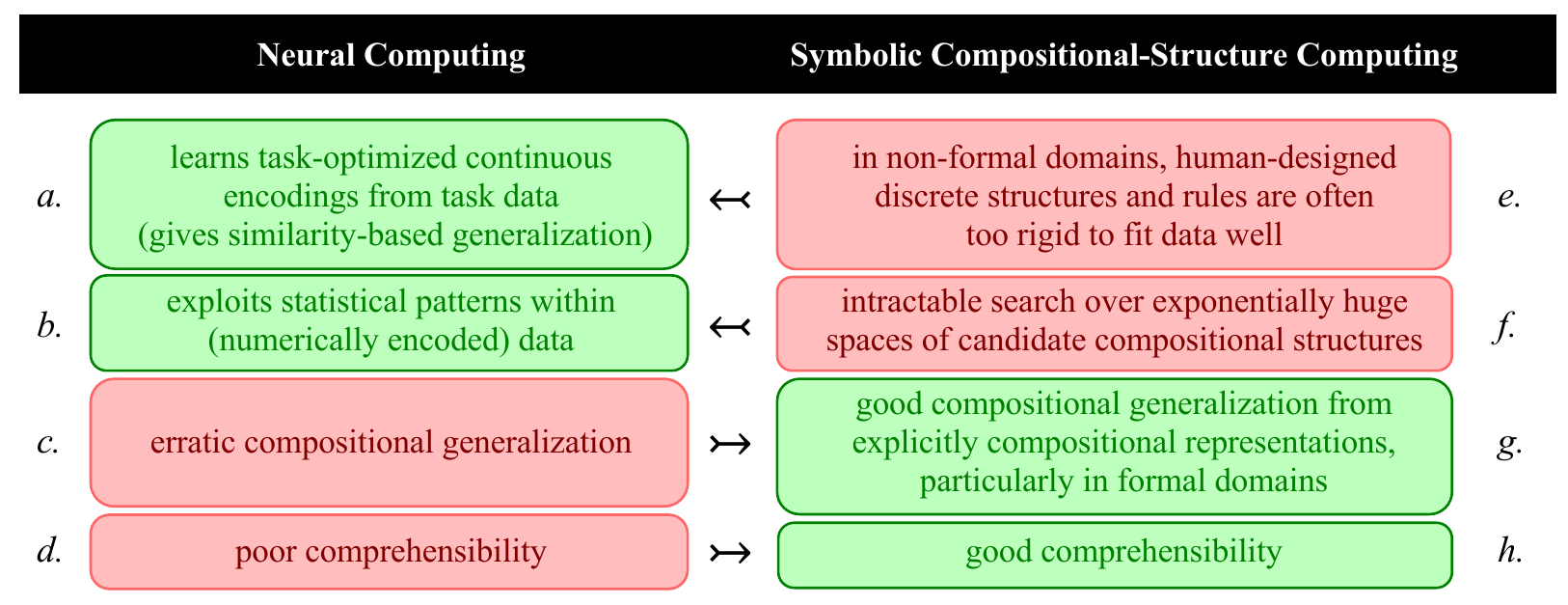} 
    \centering
    \caption{Neural and symbolic compositional-structure computing afford complementary strengths.}
    \label{fig:1C}
\end{figure}
\vspace{4mm}

\squeeze{0.95}{Both symbolic- and neural-based 20th-century AI systems} were non-neurocomposition-al because they violated either the Continuity or the Compositionality Principle---resulting in serious deficiencies.
Human intelligence is possible because it deploys a type of computing that simultaneously respects both principles: neurocompositional computing.

\clearpage

\scisec{\squeeze{0.94}{21st-century AI successes: First-generation (\Gi) neurocompositional computing}}
The unprecedented surge in machine intelligence we have witnessed in the 21st century is due largely to progress in AI systems exploiting neural computing.
The field has been growing exponentially, and the diversity of network architectures is enormous.
However, we can gain insight into the roots of the progress by focusing on two of the most successful and widely used types of networks:
Convolutional Neural Networks (CNNs) and Transformers.

CNNs \citep{fukushima1982neocognitron,lecun1998gradient} have been enormously important in systems operating on visual input.
Inspired by the neural structure of mammalian visual systems, 
CNNs process an input image by analyzing patches of the image with feature detectors.
There are multiple layers of feature detectors, each computed by a set of neurons with a preferred stimulus implicitly encoded in its weights. 
Taking advantage of the Continuity Principle, a CNN's filters are learned by slowly tuning the continuous connection weights that implement them.
Each layer is structured as a 2-d grid that mirrors the 2-d spatial structure of the image patches being processed.
Within a given layer, each feature detector analyzes all patches; processing is, by design, position-invariant. 

Critically, CNN processing also incorporates the Compositionality Principle using spatial structure.
At each layer, the analysis of the whole image is gotten by composing together---spatially arranging---analyses of larger patches of the previous layer's analysis of its smaller patches.

Another, more recent, architecture for neural computing, the Transformer \citep{vaswani2017attention}, has given rise to a quantum leap in machine language processing \citep{qiu2020pre}.
A Transformer takes an entire sequence of symbols as input and produces a sequence of symbols as output.
To compute the activation vector for a given symbol at a given layer, the Transformer calculates a weighted sum of information from vectors encoding other symbols.
It can be shown that this is equivalent to a graph with weighted links between symbols: links along which data flows. In state-of-the-art models, these links partially align with abstract inter-word relations posited in linguistic theory \citep{manning2020emergent}.

Thus both of these truly seminal network types---CNNs and Transformers---derive much of their power from their additional compositional structure \citep{henderson2020unstoppable}: spatial structure and a type of graph structure. However, these \eB{first-generation (\Gi) 
neurocompositional computing} architectures are insufficient for general higher cognition (e.g., reasoning,  language). In these domains, compositional structure is not physical but abstract, like a planning structure in which plans contain sub-plans which contain sub-sub-plans \citep{hendler1990ai}. CNNs, limited to spatial structure, are clearly unable to capture such structure.

Transformers, on the other hand, use graphs, which in principle can encode general, abstract structure, including webs of inter-related concepts and facts. However, in Transformers, a layer's graph is defined by its data flow, yet this data flow cannot be accessed by the rest of the network---once a given layer's data-flow graph has been used by that layer, the graph disappears. 
For the graph to be a bona fide encoding, carrying information to the rest of the network, it would need to be represented with an activation vector that encodes the graph's abstract, compositionally-structured internal information. 
The technique we introduce next---NECST computing---provides exactly this type of activation vector.

\scisec{Deepening neurocompositionality through a solution to the Central Paradox: \\ NECST computing}
\Gi\ neurocompositional computing, illustrated by the CNN and Transformer architectures, has contributed greatly to the progress of AI in the 21st century.
The work we now present pursues neurocompositional computing more aggressively, aiming for stronger incorporation of the Compositionality Principle by instilling network capabilities that enable explicit construction and processing of general, abstract, compositionally-structured activation-vector encodings---while staying strictly within the confines of neural computing to respect the Continuity Principle.
This is \e{\Gii\ neurocompositional computing}.

\scisubsec{Neurocompositionality and compositional generalization.} Increasingly deep implementation of neurocompositionality brings increasingly robust compositional generalization.
To provide a straightforward initial example, we consider
an extremely simple task: taking as input a sequence of five digits (e.g., $\langle$\Nfont{3,9,7,4,7}$\rangle$), encoding the entire sequence internally, and then reproducing that sequence as the output (Fig.~4).
This copying task demands compositional generalization because it requires learning that if the digit \Nfont{4} appears in position \textit{n} in the input, then \Nfont{4} must be placed in position \textit{n} of the output, for every possible position \textit{n}.
Similarly, if the digit \Nfont{n} appears in position \textit{4} in the input, then \Nfont{n} must be placed in position \textit{4} of the output, for every possible digit \Nfont{n}.

We test such compositional generalization by withholding from training all sequences of a certain type: `\NinN' sequences, in which there is a \Nfont{1} in position \textit{1}, or a \Nfont{2} in position \textit{2}, etc.
After training, correctly copying the unseen \NinN\  sequences requires compositional generalization because digit \Nfont{n} has never been seen in position \textit{n} before.

To see how increasing the degree of neurocompositionality improves compositional generalization, we train a
sequence of successively more neurocompositional networks on the copying task (withholding \NinN\ examples):
\textsc{CopyNet-0}, a pre-Transformer (long-short-term memory, LSTM) neural network \citep{hochreiter1997long} which has little built-in neurocompositional structure; \textsc{CopyNet-1G}, a Transformer, which uses a mildly neurocompositional data-flow-graph structure (\Gi\ neurocompositional computing); and \mbox{\textsc{CopyNet-2G}}, a more thoroughly neurocompositional NECST model (\Gii\ neurocompositional computing: a NECSTransformer, described below).

\begin{figure}[!ht]
\begin{subfigure}[b]{0.04\textwidth}
    \textit{a}.\newline\newline\newline\newline\newline\newline\newline\newline
\end{subfigure} %
\begin{subfigure}[t]{0.43\textwidth}
    \centering
    \includegraphics[width=\textwidth]{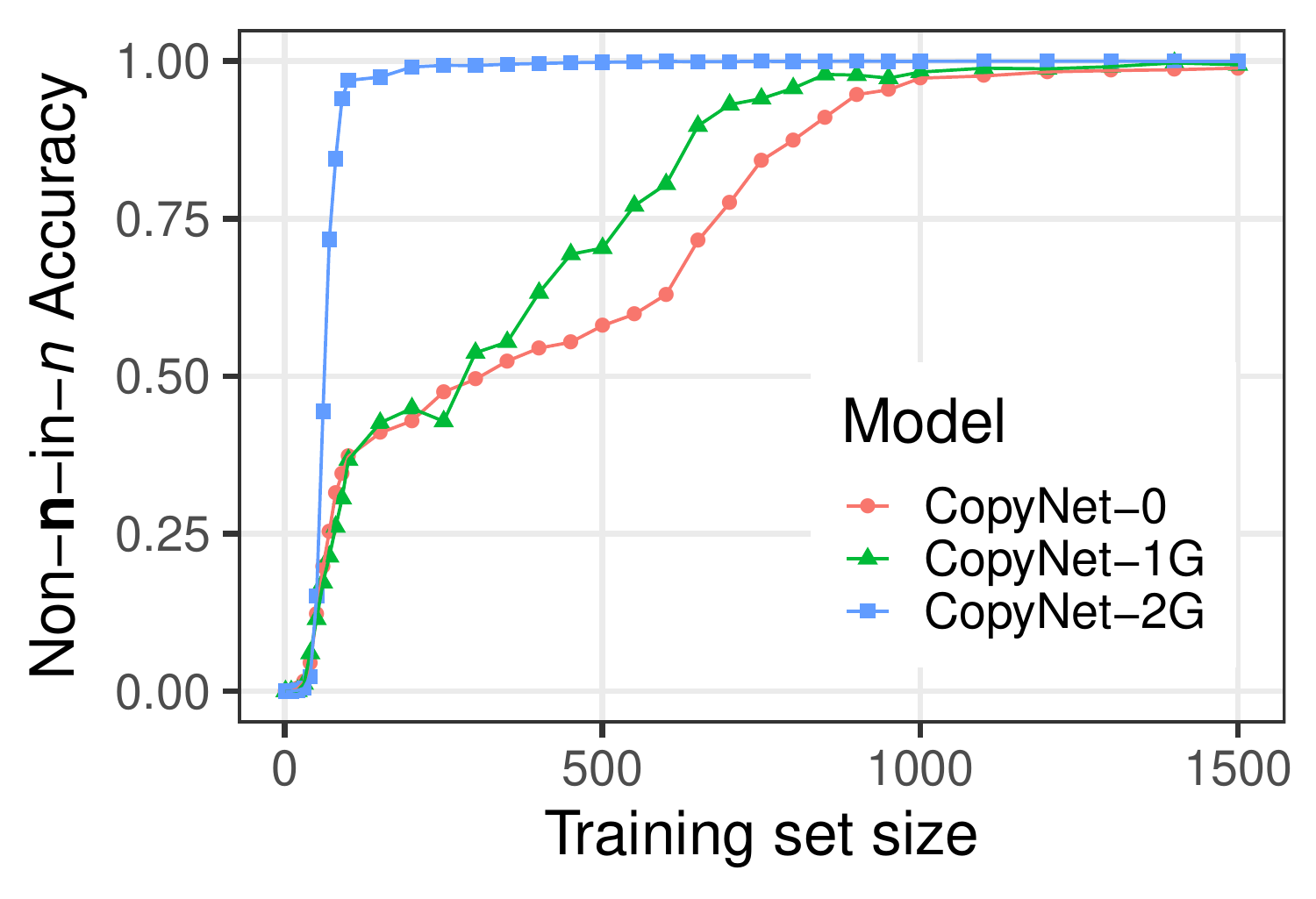}
\end{subfigure} %
\hfill
\begin{subfigure}[b]{0.04\textwidth}
    \textit{b}.\newline\newline\newline\newline\newline\newline\newline\newline
\end{subfigure} %
\begin{subfigure}[t]{0.43\textwidth}
    \centering
    \includegraphics[width=\textwidth]{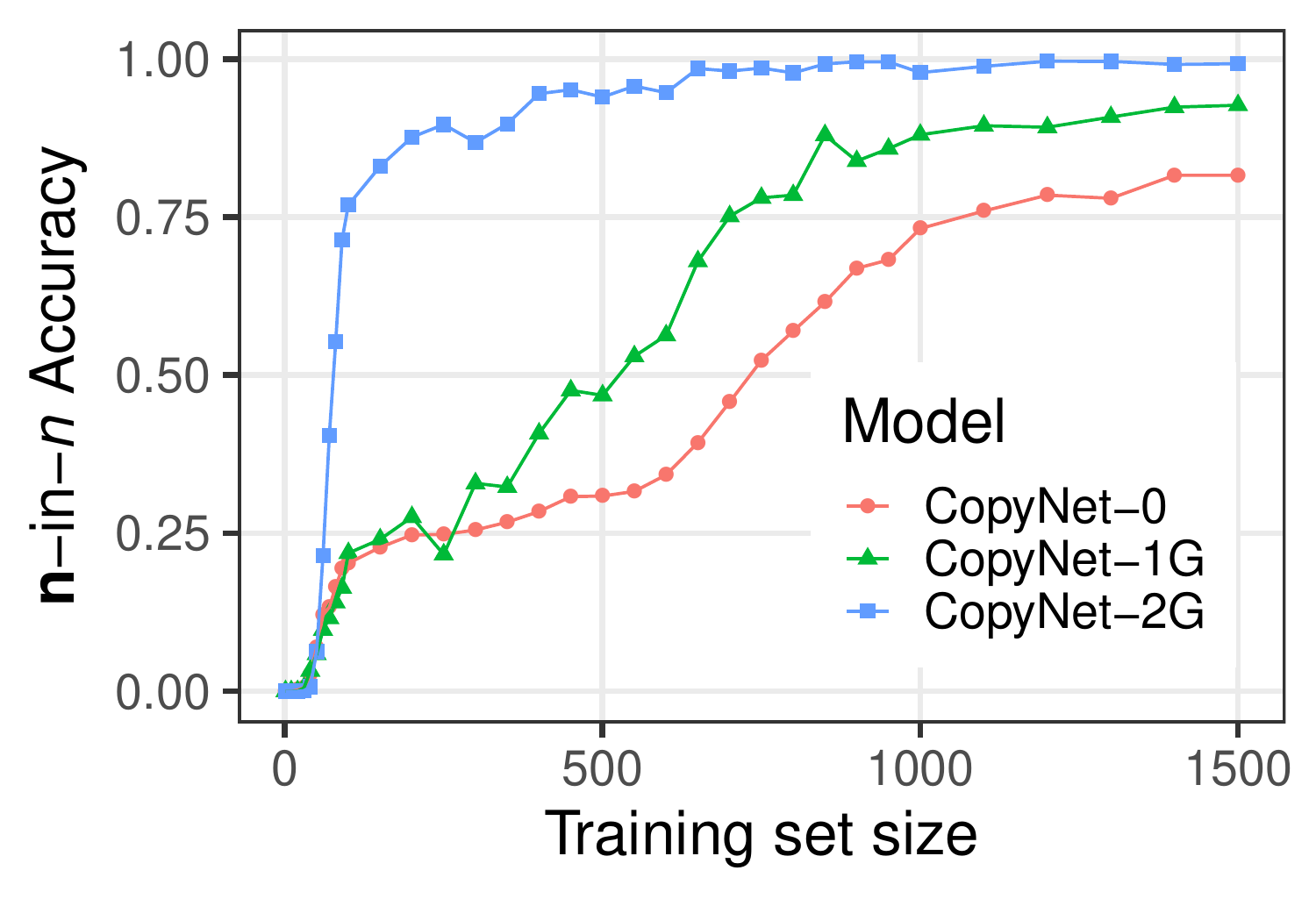}
\end{subfigure} %
\begin{subfigure}[b]{0.04\textwidth}
    \textit{c}.\newline\newline\newline\newline\newline\newline\newline\newline
\end{subfigure} %
\begin{subfigure}[b]{0.43\textwidth}
    \centering
    \includegraphics[width=\textwidth]{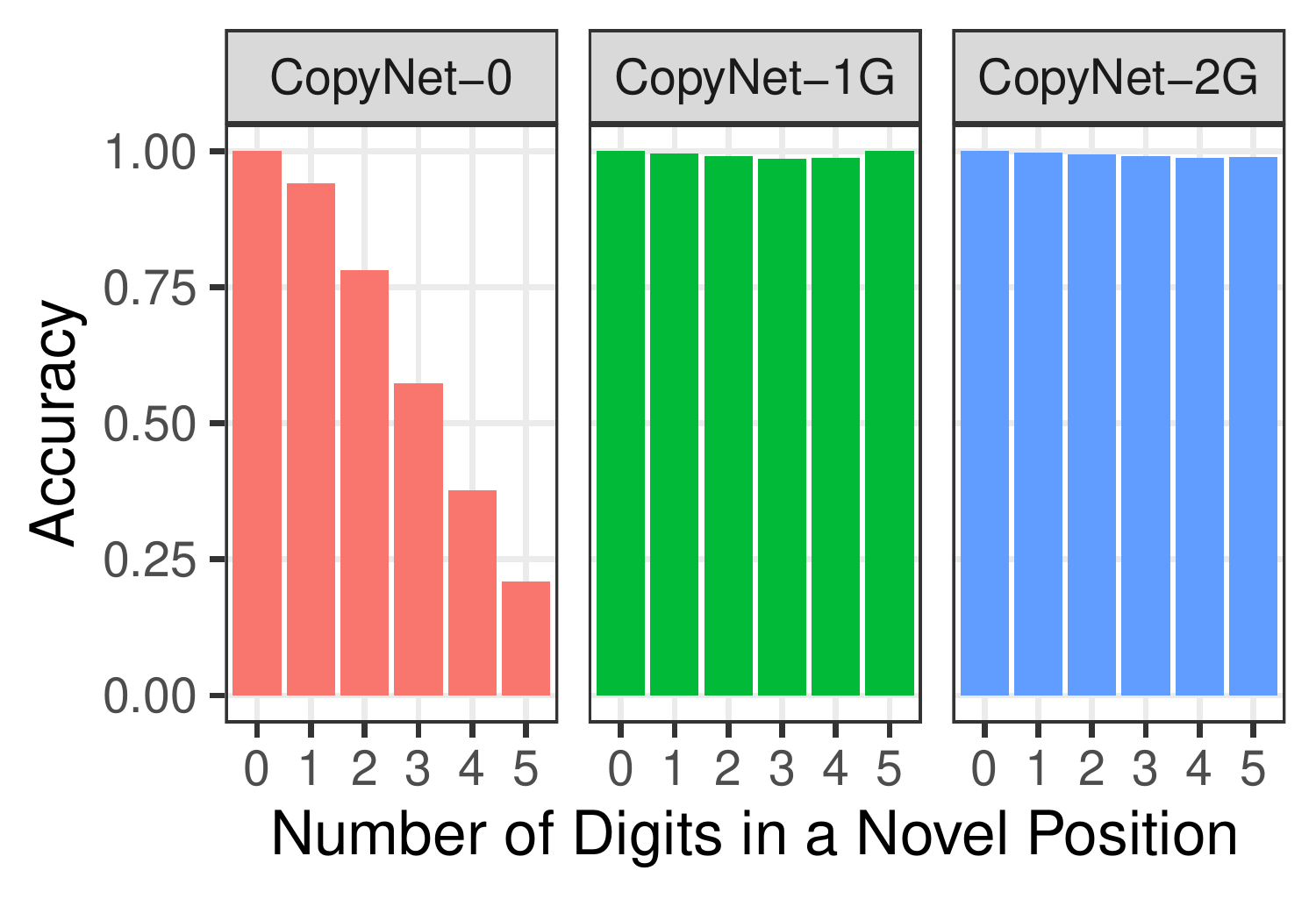}
\end{subfigure} %
\hfill
\begin{subfigure}[b]{0.04\textwidth}
    \textit{d}.\newline\newline\newline\newline\newline\newline\newline\newline
\end{subfigure} %
\begin{subfigure}[b]{0.43\textwidth}
    \centering
    \includegraphics[width=\textwidth]{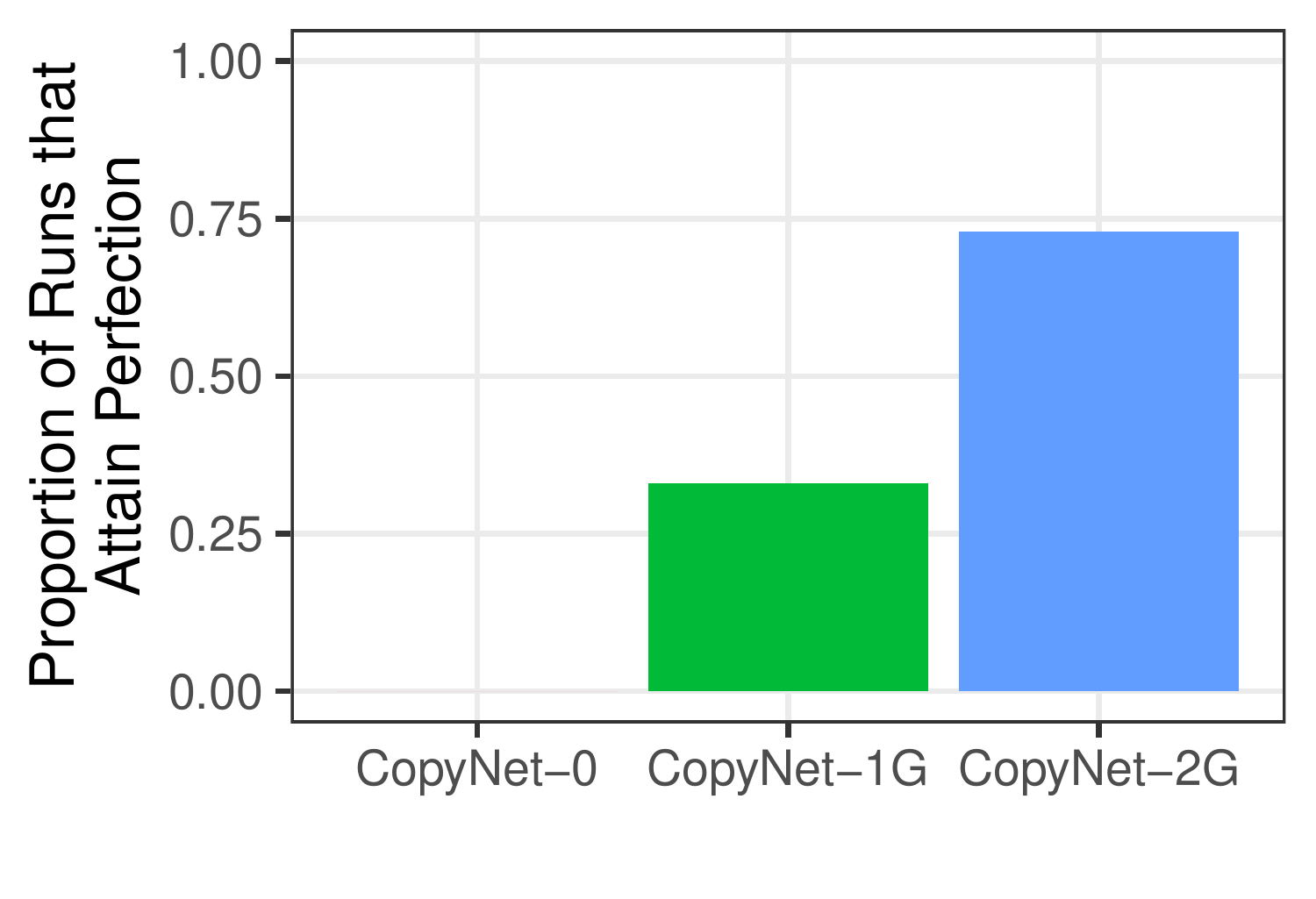}
\end{subfigure} %
    \centering
    \caption{Successive increases in a model's degree of neurocompositional structure bring improvements in compositional generalization.}
    \label{fig:copynet}
\end{figure}

Fig.~4 gives several illustrations of the improved compositional generalization that comes with increased neurocompositional structure.
Figs.~4\textit{a–b} show how models with greater degrees of neurocompositionality perform better with less data---in Fig.~4\textit{b}, when compositional generalization is required, and in Fig.~4\textit{a} when it is not.
In the latter case, all models eventually reach 100\% correct, but we see that \sc{CopyNet}-2G reaches this with an order of magnitude fewer training examples. 
Fig.~4\textit{c} shows that \sc{CopyNet}-0 does progressively worse on sequences with more and more digits in positions they never occupied during training, while \sc{CopyNet}-1G and \sc{CopyNet}-2G show little degradation in performance.
Given an \NinN\ input, e.g., $\langle$\Nfont{3,9,7,4,7}$\rangle$ (with \Nfont{4} in position \textit{4}), \sc{CopyNet}-0 is prone to produce an erroneous output in which neighboring digits are flipped to move \Nfont{n} out of position \textit{n}, e.g., $\langle$\Nfont{3,9,4,7,7}$\rangle$ (with \Nfont{4} now in position \textit{3}).
In contrast, \sc{CopyNet}-1G and \sc{CopyNet}-2G have no difficulty performing the compositional generalization required to produce \NinN\ outputs.

How do these neural learners’ final performance levels compare to a symbolic learner, which would be expected to reach perfect accuracy?
We trained 100 distinct instances of each type of model and measured how many of the 100 instances learned the task perfectly (scoring 100\% on the set of all possible digit sequences of length 5—including all \NinN\  sequences). 
As Fig.~4\textit{d} shows, no \sc{CopyNet}-0 instance performed perfectly; about one third of \sc{CopyNet}-1G instances reached perfect performance; and over two thirds of \sc{CopyNet}-2G instances attained perfection.

In sum, greater neurocompositionality brings faster learning (Fig.~4\textit{a--b}), more robust compositional generalization (Fig.~4\textit{c}), and an increased chance of learning the task perfectly (Fig.~4\textit{d}). 
All of these improvements lie in areas that have traditionally been strengths of symbolic systems yet weaknesses of neural systems (as exhibited by \sc{CopyNet}-0).
These results, along with findings on more complex tasks reported below, support the conclusion that neurocompositionality endows neural models with some of the strengths of symbolic models while retaining the power of deep learning. 

So how, exactly, has neurocompositional computing progressed to \Gii?

\scisubsec{NECST computing defined.} At the foundation of \Gii\ neurocompositional computing is the question: 
How can fully general, abstract compositional structures be encoded in neural activation vectors so that composi\-tional-structure processing can be carried out on these encodings using only continuous neural computing? The answer we adopt from cognitive science---NECST \citep{smolensky2006harmonic}---is a solution to the Central Paradox of Cognition: 
How can our own cognitive encodings simultaneously be neural encodings and compositional encodings \citep{hofstadter1979godel}?

The first insight exploited by NECST is that buried within compositional structures are two distinct types of information that are tightly bound together: \textit{what} information and \textit{where} information.
Crucially, \textit{where} refers to a position in an abstract structure, not literally a position in physical space.

Consider a simple, discrete compositional structure: a sequence of two symbols, i.e., an ordered pair.
In \lockable, the \textit{what} comprises two symbols,  $\sf{lock}$ and $\sf{able}$; the \textit{where} comprises two abstract positions: $\textit{\sf{L}}$ (left) and $\textit{\sf{R}}$ (right). 
\Lr\ and \Rr\ are \textit{structural roles}, and they define a structural type: the ordered pair.
We can visualize \textit{\sf{L}} as \sf{$\left[ \ \boxed{{\color{white} x}} \  \textrm{---} \  \right]$} 
and \textit{\sf{R}} as \sf{$\left[ \  \textrm{---} \  \boxed{{\color{white} x}} \  \right]$}.
In [\sf{lock} \sf{able}], role \textit{\sf{L}} is filled by \sf{lock} and role \textit{\sf{R}} is filled by \sf{able}.
This structure comprises two \textit{filler-role bindings}, each of which we write as \sf{filler}\textit{:\sf{role}}.
Our pair can be written as \lock\textit{:}\Lr\ \textit{\&} \able\textit{:}\Rr\ 
and visualized as  \sf{$\left[ \ \boxed{\sf{lock}} \  \textrm{---} \  \right]$ \textit{\&} $\left[ \  \textrm{---} \  \boxed{\sf{able}} \  \right]$};
the two bindings are aggregated (by the \textit{\&} operator) to form a single structure.
In NECST computing, this formal characterization of the structure is neurally encoded as a kind of vector, as follows.

Each filler symbol is encoded by an activity vector, a list of real numbers (potentially resulting from deep learning).
This is commonplace in neural network modeling: as previously observed, replacing discrete symbols with continuous vectors is at the heart of the 21st-century revolution in machine language processing.
A key innovation of NECST is that each \textit{structural role} is similarly encoded by an activity vector.
Then the two operations, \sf{:} for binding each filler to its role and \sf{\&} for aggregating the bindings, are realized as operations on vectors.
The aggregation operation \sf{\&} becomes the summation of vectors, \textsl{+}.
The binding operation \sf{:} is a little more complex; it becomes the tensor product $\otimes$ (possibly compressed).%
\footnote{The tensor product yields encodings that are a special kind of vector: a tensor, which we can take to be a set of real numbers that can be visualized as forming a multi-dimensional grid.
For the tensor product of a filler vector $\mathbf{f} = (f_1, f_2, \mathellipsis)$ and a role vector $\mathbf{r} = (r_1, r_2, \mathellipsis)$, this is a 2-D grid; the number in the $i^\textrm{th}$ row and $j^\textrm{th}$ column is simply $f_i r_j$.}

The resulting neural encoding is called the \e {Tensor Product Representation} (TPR) of the compositional structure \citep{smolensky1990tensor}.
Hence the name, \e{Neurally-Encoded Compositionally-Structured Tensor} (NECST) computing.
There are several neurocompositional architectures known by different names \citep{gayler2003vector,eliasmith2013build,garcez2019neural}, but many of these turn out to be based on TPRs that have been compressed by a linear transformation \citep[Ch. 7]{smolensky2006harmonic}.

Crucial to the power of symbol structures is that symbols maintain their identity as they play different roles (\citealp[p. 1394]{mitchell2010composition}).
The role \Rr\ of \unlock\ and the role \Lr\ of \lockable\ are filled by the same symbol, \lock; in a symbolic language-processing model, this is why \textit{lock} contributes the same meaning to both \textit{unlock} and \textit{lockable}---making compositional generalization possible.
This critical property also holds for the continuous, explicitly compositional vector encodings provided by TPRs, because the same vector that encodes \lock\ appears in the vectors that encode the binding \sf{lock}:\Rr\ and the binding \sf{lock}:\Lr.
TPRs successfully disentangle the \textit{what} and the \textit{where} information carried by compositional structure; these very different types of information are thoroughly fused in traditional non-compositional neural models.

All types of compositional structures comprise a set of \sf{filler:}\textit{\sf{role}} bindings \citep{newell1980physical}: TPRs provide a fully general method for neurally encoding compositional structure. 
Further, the mathematical operations that underlie TPRs enable systematic, compositional generalization, so that TPRs do not just encode compositional structure but specifically do so in a way that facilitates the behavioral benefits of compositional structure. 
In fact, mathematical results show that with inputs and outputs encoded as TPRs, neural networks can be designed to precisely compute many complex compositional-structure processing functions---including the types of functions shown in Fig.~2 \citep{smolensky2012symbolic}.
Thus the intricate structure-sensitive processing that empowers symbolic computing \citep{fodor1988connectionism} can be performed by networks using TPR encodings.
In cognitive science, theories of language based in TPRs have given rise to new computational architectures for grammars; these have significantly advanced linguistic theories explaining the compositional structure of language \citep{pater2009weighted,prince1997optimality}.
Networks processing TPRs can enforce the strong kind of constraints on compositional processes that are imposed by grammars.

Thus NECST does indeed provide an answer to the Central Paradox: it provably allows neural computing to carry out important types of compositional-structure processing previously only computable by discrete symbolic computing.
But NECST goes beyond symbolic computing, reaping the benefits of the Continuity Principle.
TPRs provide a new type of encoding: \bi{continuous compositional structure} \citep{goldrick2016coactivation,smolensky2014optimization}.
Returning to the example of the ordered pair structure, halfway between the two vectors encoding the discrete roles \Lr\ and \Rr\ is a vector which encodes a continuous role; this \textit{blend} of the \Lr\ and \Rr\ roles can be bound to a filler such as \sf{t} to produce the TPR for a binding we can visualize as 
\sf{$\left[ \ \boxed{\mbox{\scriptsize \ \ \ \textsf{t} \ \ \ }} \  \right]$}.
Here, \sf{t} fills a weak, half-strength role that spans both the left and right halves of the pair: \sf{t} is half-present in both places at once. 
Moreover, TPRs enable a single role like \Lr\ to be filled with a blend of symbols, each simultaneously present (or active) to a continuously variable degree.

Such continuous structures may seem odd, but recent work in linguistics \citep[e.g., ][]{rosen2019learning,zimmermann2019gradient} has shown how they make possible theories that cover bodies of data that no single theory using discrete structure can.
For example, basic speech sounds can fill continuous roles: the \sf{t} sound appearing in the middle of the French phrase \textit{peti\textbf{\underline{t}} ami} (literally, `small friend') simultaneously resides both at the end of the first and the beginning of the second word \citep{smolensky2020learning}: exactly a case of 
\sf{$\left[ \ \boxed{\mbox{\scriptsize \ \ \ \textsf{t} \ \ \ }} \  \right]$}.
Such continuous compositional structure can explain observed complex sound patterns that can be explained by no single theory---even a probabilistic one---that forces such sounds into discrete roles.

\scisec{Progress from early NECST AI: \Gii\ neurocompositional computing}
To illustrate \Gii\ neurocomputing, we present the \TPT\ \citep{schlag2019enhancing}.
Recall that the original Transformer constructs its own implicit data-flow graph for each input symbol sequence, for each layer; this determines, for a given symbol, the extent to which other symbols are consulted for information to determine the vector that encodes the given symbol at that layer.
In the \TPT, this vector has explicit compositional structure: it is the TPR for a continuous structure composed from a given number---$n_r$---of parts determined by $n_r$ roles, roles which vary across symbols, layers, and inputs, and result from deep learning.
To create its TPR at a given layer, each symbol generates $n_r$ vectors to encode its roles, and computes a filler for each of these roles.
Each filler is a weighted sum of encodings of other symbols in the same layer. 

This model generates a graph at each layer, but goes beyond the plain Transformer by labeling each link with a vector that encodes a relation: a relation that holds between the symbol that sends a filler and the  symbol receiving it, which binds it to the corresponding role vector it has generated. 
Crucially, this relationship is no longer merely implicit in the information flow: it is inserted explicitly into the TPR activation vector that encodes the symbol, making it information that is fed directly to subsequent layers for further processing.
For example, in a math-problem-solving application, the TPR created to encode a digit in a divisor includes a role interpretable as `denominator', bound to a filler sent from a numerator digit.

\scisubsec{Interpreting structural relations invented through deep learning.}
The explicit encoding of relations computed by the \TPT\ enables it to learn general abstract graph structures; in fact, each  \TPT\ model uses deep learning to invent its own type of compositional structure to optimally perform its task. On math problem solving, when encoding a ratio of ratios, the invented structures implicitly exploit the inference rule in Fig.~2\textit{b}, assigning the same role to the denominator-of-a-denominator as to a numerator-of-a-numerator (respectively \textit{v} and \textit{x} in Fig.~2\textit{b}) \citep{schlag2019enhancing}. 

Two other NECST models that process English invent structures that are partially interpretable in grammatical terms, although no information is given to the models about grammar in general or the structure of English in particular.
In a model that learns to answer questions about Wikipedia articles, \sc{QANet}, the roles learned for TPR encodings of English questions partially align with recognizable linguistic structural properties, at levels ranging from small features of words (such as \sc{plural}), to general semantic categories like \sc{predicate} (expressed by verbs and adjectives) to an invented type of multi-word sequence that could be called a \textit{wh}-\sc{restrictor-phrase}, like \textit{what famous event in history} \citep{palangi2018question}.
Another model, \sc{CaptionNet}, learns to generate image captions, in the process inventing roles that are highly predictive of the sequence of parts of speech (noun, adjective, etc.) in the captions it generates \citep{huang2018tensor,huang2019attentive}.

\scisubsec{Improved compositional generalization.}
\sc{StoryNet} \citep{schlag2018learning} is a NECST model that answers questions about short synthetic narratives \citep{weston2015towards}, such as the one shown in Fig.~5\textit{b}.
To test compositional generalization, during training certain character names were allowed to appear in only some fraction of the multiple subtasks in this dataset; the model was then tested on all subtasks using these withheld character names.
The smaller the fraction of subtasks in which a given name was seen during training, the more compositional generalization is required, and as the plot shows, the more \sc{StoryNet} decisively dominates the previous state-of-the-art system.

On two highly challenging natural tasks involving much more than simple compositionality, NECST models achieved new state-of-the-art performance.
One is \sc{CaptionNet}: image captioning requires learning an extremely complex relationship between the two very different, approximately compositional, natures of images and sentences. 
Another is a NECSTransformer model that generates text summaries \citep{jiang2021enriching}, \sc{SummaryNet}.
Without being told to do so, the model invents structure that parcels syntax---sentence structure---into roles, and semantics---meaning---into fillers.

\begin{figure}[!ht]
    \centering
    \includegraphics[width=\textwidth]{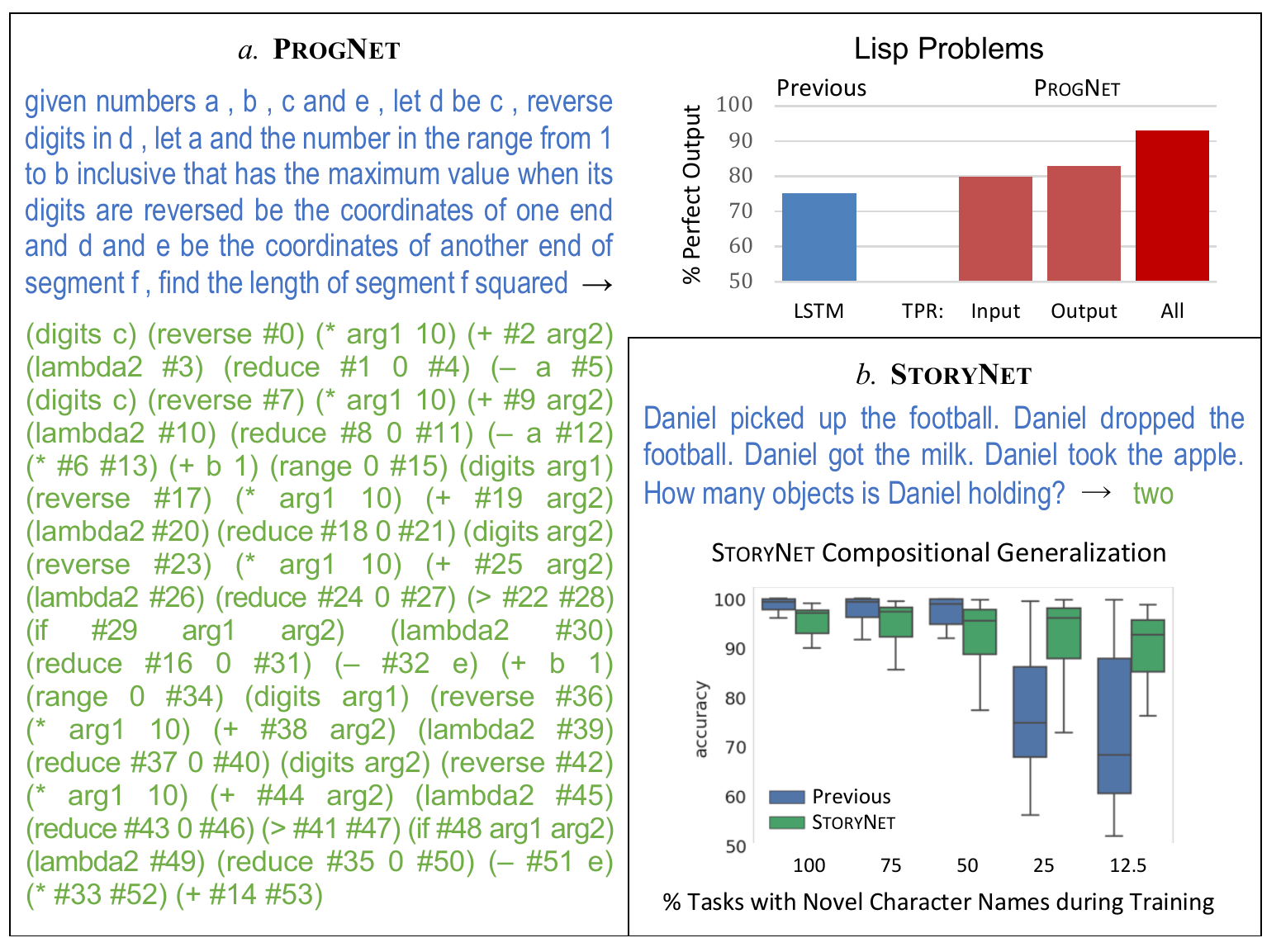}
    \centering
    \caption{
    NECST models showing  improved compositional generalization over previous models from designed, symbolically-motivated inductive biases---\sc{ProgNet}: \citet{chen2020mapping}; \sc{StoryNet}: \citet{schlag2018learning}.
    }
    \label{fig:4}
\end{figure}

\scisec{Benefits of improved comprehensibility from \Gii\ neurocompositional \\ computing}
In addition to improved compositional generalization, closer alignment with the type of computing underlying human cognition renders NECST AI systems more human-comprehensible, as already illustrated by \sc{QANet} and \sc{CaptionNet}.
Because \Gii\ neurocompositional AI systems are significantly less opaque than previous neural models, they deliver several important benefits for overcoming serious problems facing the development of \squeeze{0.93}{more explainable, trustworthy, and controllable AI systems \citep{adadi2018peeking}}. 

\scisubsec{Informing deep structure learning.}
Centuries of successful analysis of cognition as compositional-structure processing have given cognitive science and traditional symbolic AI powerful symbolic theories for computing cognitive functions, but the computational chasm separating symbolic compositional processing from traditional non-compositional neural computing has rendered this wealth of insight largely unusable for constructively informing the development of contemporary, deep-learning-based AI.
Neurocompositional computing opens the door to using such theories to inform---i.e., improve the inductive bias of---neural learning. 

An example is provided by the NECST-generation AI model, \sc{ProgNet} (Fig.~5\textit{a}), which learns to take a problem stated in English and produce a computer program that solves the problem; one version solves math word problems, another, Lisp-programming problems \citep{chen2020mapping}.
Like \sc{CaptionNet} and \sc{SummaryNet} above, \sc{ProgNet} solves a task requiring far more than identifying compositional structure. Computing the linguistic subparts of the English problem statement (e.g., in Fig.~5\textit{a}, \sf{the number in the range from 1 to b inclusive that has the maximum value when its digits  are  reversed}) is an important first step, but translating such English expressions into the corresponding sequence of Lisp commands requires rich knowledge of how the syntax and semantics of English expressions correspond to the syntax and semantics of Lisp operations, and knowledge of how composition of the English subparts relates to composition of the Lisp subparts.

Previous models for such program-generation tasks produce the output program one symbol at a time, but NECST enables models that take advantage of the structure inherent in the programs: a sequence of commands, each a tuple consisting of an operation and the arguments it operates upon, such as \sf{(add,~x,~5)} or \sf{(append,~list1,~list2)}.
\sc{ProgNet} is designed to bias learning towards producing not a sequence of individual symbols, but a sequence of TPRs, each decodable as such a tuple of symbols comprising a command.
Informing the model in this way about the structure of its outputs allowed it to set a new state of the art for both problem-solving tasks, generating Lisp programs as long as 55 commands  perfectly.
And the similarity structure of the learned vector encodings of symbols can be partially interpreted: for example, arithmetic operators cluster together, as do geometric functions like \sf{area}, and operations for processing character sequences.

A second example, already introduced, is \sc{StoryNet} \citep{schlag2018learning}, which was provided with valuable biases for compositional learning (recall Fig.~5\textit{b}).
This network learned to encode abstractions of entities and relations as vectors and to use built-in operations to bind them together into a TPR that explicitly encodes a continuous graph structure capturing knowledge of the events in the narrative.
The learning bias provided to this network was a set of useful built-in operations for updating, as each sentence of the narrative arrives, the vector encoding the continuous knowledge graph, and for sequentially extracting information from the graph to answer questions.
The model successfully learned how to use these operations.

\scisubsec{Diagnosing errors.}
Recall that the NECST model \sc{QANet}, in the service of learning to answer questions about Wikipedia articles, invents a number of roles that can be interpreted grammatically.
In addition, some learned fillers can be interpreted semantically (Fig.~6\textit{b}).
The word \sf{Who} has several meanings, and which filler is assigned to a particular instance of \sf{Who} correlates with which meaning is appropriate for the specific context of that instance. 
In the articles' many mentions of the TV character named \sf{Dr. Who}, the model sometimes assigns \sf{Who} the filler that would be appropriate for a question word, as in \sf{Who died?} (although \sc{QANet} made this error at a considerably lower rate than a state-of-the-art probabilistic symbolic system designed for analyzing English sentences, \citealt{manning2017stanford}).
When the model made this internal filler error, it was 5 times more likely to produce an incorrect output---an output that would have been appropriate had \sf{Who} actually been a question word rather than a character name \citep{palangi2018question}.

\begin{figure}[!ht]
    \centering
    \includegraphics[width=\textwidth]{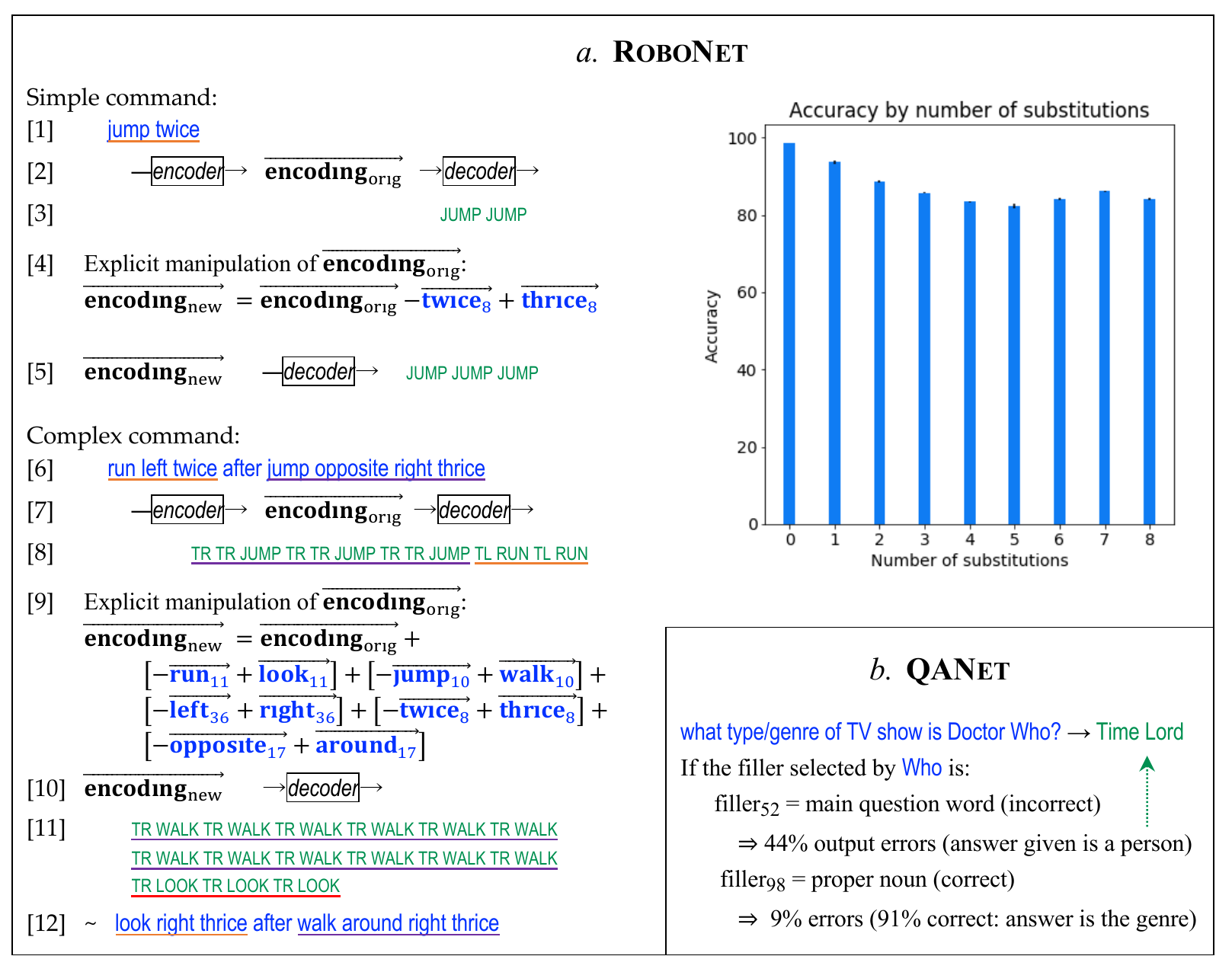}
    \centering
    \caption{Example NECST analyses demonstrating new capabilities enabled by greater comprehensibility: Controlling output and diagnosing errors---\sc{RoboNet}: \citet{soulos2020discovering}; QANet: \citet{palangi2018question}.}
    \label{fig:5}
\end{figure}

\scisubsec{Controlling output.}
The incomprehensibility of contemporary black-box non-composi\-tion\-al neural AI models severely limits our ability to control them. This lack of control is problematic; for example, it makes it challenging to address the socially toxic biases the outputs of these models often display \citep{bender2021dangers}.
In contrast, the relative comprehensibility of NECST encoding vectors
can enable direct intervention---precision surgery---on internal activation patterns in order to control the output of a network.

Our illustration of this introduces a discovery revealed by NECST analysis techniques about the internal encodings learned by standard neural networks \citep{mccoy2019rnns}. (For a demo, see \url{http://rtmccoy.com/tpdn/tpr\_demo.html}.) 
One case study concerned a simple hypothetical robot: given an input description of a maneuver in highly simplified English, the task of the network (which we dub \sc{RoboNet}) is to produce an appropriate output sequence of basic commands to carry out that maneuver \citep{lake2018generalization}.

When \sc{RoboNet} is given extensive training data that widely exemplifies the compositional possibilities inherent in its highly compositional artificial task, it achieves good compositional generalization.
A NECST-based analysis revealed that in this case
the key internal encoding can be extremely well approximated by a TPR encoding of the input, with appropriate role assignments to input words and appropriate vector encodings of these words and their roles \citep{soulos2020discovering}.
As depicted in Fig.~6\textit{a}, this understanding of \sc{RoboNet}'s learned encodings as TPRs of approximately-discrete structures allows us to write a closed-form expression for the learned vector [2] that internally encodes any given input [1], which in turn tells us exactly how to take the internal encoding of an input such as \sf{jump twice} and directly modify it into the encoding that would be produced by \sf{jump thrice}: we subtract the vector hidden in this encoding that encodes \sf{twice} and add the vector that encodes \sf{thrice} [4]---in the process, changing the activation level of every neuron by a precisely-determined amount.
Now the output of the model changes from \textsf{\color{ForestGreen}\sc{jump jump}\color{black}}\ [3] to \textsf{\color{ForestGreen}\sc{jump jump jump}\color{black}}\  [5]; that is, the alterations that we have made to the model's internal encodings produce exactly the behavioral alteration we intended to make, illustrating how the interpretable structure of TPRs facilitates control of a model's behavior.

Considerably more complex modifications, and extended sequences of successive modifications, can also be successfully effected in this way.
The eight-word input in [6] produces the encoding vector [7] and output [8].
We can change this encoding to match that of a virtual input that differs from it in five words [12] by taking five steps of subtracting the encoding of an existing word and adding in the encoding of a new virtual word [9].
(\sf{TL}, \sf{TR} denote Turn-Left and Turn-Right; $\vec{\sf{twice}_\sf{8}}$ is the vector encoding of \sf{twice}:\textit{\sf{8}}, the binding of filler \sf{twice} to role \textit{\sf{8}}.)
The resulting encoding produces the output [11] that correctly corresponds to the virtual new input [12]: the original 13 output commands have been replaced by 30 new ones.
The plot shows how, even after many such successive substitutions, the internal modifications specified by our NECST analysis continue to control behavior with high accuracy.

\scisec{Current limitations: The road to \Giii\ neurocompositional computing}
The deployment of TPRs in AI has already led to significant progress, but the gains are clearly still short of those expected from full compositional abilities.
Why is that?
The work to date is only the first step towards fully achieving NECST computing, falling short in several respects.
Identifying these shortcomings, and proposing techniques for addressing them, provides a road to \e{\Giii\ neurocompositional computing}. 

First, while the TPRs of \Gii\ systems successfully disentangle the \textit{what} and \textit{where} aspects of compositional structure, there is nothing to explicitly bias the model to learn processing that takes advantage of this disentanglement to promote compositional generalization.
Building such a bias into the network structure will be a key component of \Giii\ neurocompositional computing.

Second, the mathematics of TPRs allows them to embed inside one another just as \sf{$[ \sf{lock} \ \sf{able} ]$} can be embedded within \sf{$[ \un \ [ \sf{lock} \ \sf{able} ] ]$}. 
The corresponding neural operations are beyond 2G NECST computing, but are now in development, and will be integral to \Giii.

Third, the transformative advances in grammatical theory derived from NECST center on the innovation that grammatical expressions are encoded in activation vectors that optimally satisfy grammatical constraints which are encoded in the strengths of connections joining neurons 
(Prince and Smolensky, 1993/2004, 1997) \nocite{prince1997optimality,prince1993optimality}.
To go beyond the relatively modest gains in natural language processing observed in 2G neurocompositional computing, incorporating such optimization-based processing into \Giii\ neurocompositional computing may set the stage for major breakthroughs in language processing, as it did in linguistic theory.
An initial step in this direction shows how NECST can strengthen inference ability by enabling the meaning of symbols denoting particular entities and relations to adapt continuously to---to be optimized for---their structural context within a knowledge graph of facts \citep{lalisse2019augmenting}.

The research program reviewed here aspires to a reconvergence of AI and cognitive science through a unified theory of the computing underlying both human and machine intelligence.
Recognizing the importance of such a synergy is not new: it was already envisioned nearly two centuries ago by Ada Lovelace, who saw that in the general purpose computer, “not only the mental and the material, but the theoretical and the practical, are brought into more intimate and effective connexion with each other” \citep[Note A, p. 369]{bowden1953faster}.
It is our hope that the development of neurocompositional computing will ultimately contribute to the realization of her remarkably prescient vision of intelligent machines.

\filbreak
\bibliographystyle{aaai.bst}
\bibliography{AIMag.bib}

\section*{Acknowledgments}
We gratefully acknowledge, for support and valuable conversations, Johannes Gehrke, Li Deng, Qi Lu, Yi-Min Wang, Harry Shum, Eric Horvitz, Susan Dumais, Xiaodong He, Asl{\i} \c{C}eliky{\i}l\-maz, Chris Meek, Hamid Palangi, Qiuyuan Huang, Nebojsa Jojic, Imanol Schlag, Kezhen Chen, Shuai Tang, Laurel Brehm, Najoung Kim, Matthias Lalisse, Paul Soulos, Eric Rosen, Caitlin Smith, Coleman Haley, G\'{e}raldine Legendre, Jason Eisner, Ben Van Durme, Alan Yuille, Hynek Hermansky,
Tal Linzen, Robert Frank, J\"{u}rgen Schmidhuber, Ken Forbus, Gary Marcus, Yoshua Bengio, Steven Pinker, Jay McClelland, Alan Prince, 
Ewan Dunbar, Dapeng Wu, Randy O'Reilly, Fran\c{c}ois Charton, Guillaume Lample, Peter beim Graben, Daniel Crevier,
and all our collaborators on the papers reviewed here.
The work reported here was supported in part by NSF (GRFP 1746891, BCS-1344269, DGE-0549379) and by Microsoft Research.
Any opinions, findings, and conclusions or recommendations expressed in this material are those of the authors and do not necessarily reflect the views of the National Science Foundation or Microsoft.

\end{document}